\documentclass[conference]{ieeeconf}
\IEEEoverridecommandlockouts
\usepackage{cite}
\usepackage{amsmath,amssymb,amsfonts}
\usepackage{algorithmic}
\usepackage{graphicx}
\usepackage{textcomp}
\usepackage{xcolor}
\usepackage{multirow}
\usepackage{comment}
\usepackage{xcolor}
\usepackage{ulem}
\usepackage[absolute,overlay]{textpos}

\def\BibTeX{{\rm B\kern-.05em{\sc i\kern-.025em b}\kern-.08em
    T\kern-.1667em\lower.7ex\hbox{E}\kern-.125emX}}

\title{Wind and State Estimation on SE(3): Comparative Evaluation of EKF and UKF with Continuous and Discrete Quadrotor Models
}

\author{
Hiranya Udagedara, Adam Bigsby, Mahdis Bisheban
\thanks{Hiranya Udagedara, Adam Bigsby, and Mahdis Bisheban,  are with Dept. of Mechanical and Manufacturing Engineering, University of Calgary, Calgary, Alberta,Canada. Emails: \{hiranya.udagedara, adam.bigsby, mahdis.bisheban\}@ucalgary.ca. }
\thanks{This work was supported by the Natural Sciences and Engineering Research Council of Canada (NSERC), the Government of Alberta, Alberta Innovates, and the Schulich School of Engineering at the University of Calgary. Funding was awarded to Dr. Mahdis Bisheban, Director of the Intelligent Dynamics and Control Lab and Assistant Professor at the University of Calgary.}
}

\begin{document}

\begin{textblock*}{\textwidth}(2.0cm,0.5cm)
\centering
\small Author's preprint. Published by IEEE Conference on Control Technology and Applications (CCTA), 2026.
\end{textblock*}

\maketitle

\begin{abstract}
Use of quadrotor UAVs for wind velocity estimation is gaining popularity in recent studies, leveraging their maneuverability, compact size and low cost. Among available approaches, model-based wind velocity estimation is most commonly used, since it relies only on onboard sensors. However, as the quadrotor is a highly nonlinear system, thus making this task challenging. 
This study evaluate the use of both discrete and continuous dynamic equations of the quadrotor UAV for wind velocity estimation on SE(3), rather than commonly adapted continuous or discretized form. Lie Group Variational Integrator, developed on discrete Lagrangian is used as the discrete model without any approximation or discritization. 
The study assess both the discrete and continuous form of the quadrotor dynamics on SE(3) using Extended Kalman filter (EKF), and Unscented Kalman filter (UKF). The quadrotor UAV performance is evaluated in both MATLAB-based numerical simulations and free outdoor flight. The numerical simulations are conducted  during both hovering and trajectory-tracking flights. Results demonstrate that, by using discrete SE(3) dynamics coupled with UKF, the quadrotor achieves higher estimation accuracy while maintaining trajectory tracking, even with low-cost sensors. These findings highlight the potential of discrete quadrotor models with UKF not only for wind velocity estimation but also for other high-accuracy tasks, even when relying on low-cost onboard sensors.
\end{abstract}


\section{Introduction}
Wind velocity estimation using quadrotor UAVs, is a widely studied area of research. Use of quadrotor UAVs for such a task is being widely explored as a quadrotor UAV offers many advantages over the stationary wind collection centers and fixed-wing UAVs. These advantages are low-cost, maneuverability and compact size. 

Wind velocity measurement can be categorized to two main methods as direct measurement and model-based estimation \cite{Ahmed2024}. Studies done by Prudden et al., and Gamagedara et al., have used additional sensors such as anemometer and multi-hole pressure probe to directly measure wind velocity \cite{Prudden2018, Gamagedara2019}. However, quadrotor UAV has moving blades causing wind around the quadrotor to deviate, resulting in mutilated wind velocity measurements. Additionally, additional sensors on a quadrotor increases the payload, decreasing the flight-time. In contrast, the model-based wind estimation uses only existing sensors, and approximates the wind velocity by relying on the dynamic model of the quadrotor UAV \cite{Neumann2015}. The dynamic model of the quadrotor UAV can be obtained through several approaches. The most straight-forward approach is Euler-angle based modeling \cite{Xing2019}. However its practical use is limited due to the occurrence of  Gimbal lock. The quaternions provides a better alternative \cite{Chen2023}, but it introduces ambiguity, since the same orientation can be represented by two distinct quaternions. In contrast, modeling the system directly on the Special Euclidean group, \(SE(3)\), provides a more robust and globally consistent representation of quadrotor dynamics \cite{Shastry2021}. 

The dynamic model of the quadrotor is then used to approximate the wind velocity using measurements from the onboard sensors. These measurements are often noisy, thus estimators are used for accurate estimation. Kalman filters are the commonly used filters for the task. The studies done by Neuman et al., and Xiang et al. have used a linear Kalman filter to estimate the wind velocities, but under several constraints including hovering flight and negligible measurement noises \cite{Neumann2015, Xiang2016}. EKF on the other hand provides a better room for approximation. The EKF approximates the nonlinear quadrotor UAV system, by linearizing it around the current estimates. Therefore, several studies have used the EKF for wind velocity estimation, but under the constraints of simple flight trajectories or, constant wind conditions \cite{Gonzlez-Rocha2019, LNCSikkel2016}.  The UKF is another variant of the Kalman filter, which is nonlinear. Shastry et al. have used UKF for wind velocity estimation, while the quadrotor UAV is hovering. Though the use of EKF for wind velocity estimation is widely explored, the use of UKF is yet to be explored \cite{Shastry2023}. 

The quadrotor dynamics are often presented in continuous-time. When evaluating system performance through numerical simulations, these dynamics are often integrated using a numerical integrator such as fourth-order Runge-Kutta. Though this gives an approximate solution at very small time steps, there are residuals involve. To address this, the dynamics in this study are formulated on the Special Euclidean group \(SE(3)\), and expressed in discrete-time, enabling a more consistent and globally valid representation of position and orientation. This study aims to evaluate the effectiveness of the discrete \(SE(3)\)-based quadrotor system for wind velocity estimation, and to assess their performance relative to the commonly used continuous quadrotor system. By comparing both formulations within the EKF and UKF frameworks, the study intends to evaluate and highlight the potential advantages of discrete dynamics in improving estimation accuracy and reliability for quadrotor UAV applications.
In this paper, Section II presents the methodology, followed by the results and discussion in Section III. The paper concludes with Section IV.

\section{Methodology}
This section presents the quadrotor UAV's dynamic model, the UKF and EKF estimation processes, and implementation details.

\subsection{Dynamic Model of the Quadrotor UAV}
This section represents the continuous and discrete dynamic model of the quadrotor on \(SE(3)\). The origin of the body frame is assumed to be located at the center of mass of the quadrotor. The center of mass of the IMU is assumed to coincide with that of the quadrotor, without loss of generality. Any misalignment in orientation or displacement between the IMU and the UAV’s center of mass can be corrected through a simple IMU calibration. Once calibrated, the IMU frame remains fixed relative to the body frame, even if the body frame changes its orientation. Let \(x\in \mathbb{R}^3\) represent the position of the origin of the body frame in the inertial frame. Then \( v \in \mathbb{R}^3 \) represents the velocity of the origin of the body frame. The angular velocity of the quadrotor in the body frame is represented by \( \Omega \in \mathbb{R}^3 \), and its orientation is denoted by \( R \in \mathbb{R}^{3 \times 3} \).  \(v_w \in \mathbb{R}^3\) is the wind velocity in the inertial frame. In this study, the full state vector is \( X = [x,v,\Omega,R, v_w]\in \mathbb{R}^{21} \).

\paragraph{Continuous-Time Modeling of Quadrotor UAV}
The translational and attitude dynamics of a quadrotor in continuous-time, subject to wind disturbances are as follows \cite{chen2022incorporating, Goodarzi2017}:
\begin{align}
    \dot{x} =& v,
    \label{eq:xdot_w1}\\
        \dot{v} =& ge_3 - \frac{f_cRe_3}{m} + R\frac{D_f}{m} + \frac{\Delta_1}{m},\label{eq:vdot_w1} \\
            \dot{\Omega} =& J^{-1}(-\Omega \times J\Omega + M_{c} + \Delta_2),
    \label{eq:Wdot_w1} \\
    \dot{R} = & R\hat{\Omega},
    \label{eq:Rdot_w1}\\
    \dot{v}_w =& \mathcal{W}_w, \label{eq:vwdot_w1}
\end{align}
where \(e_3=[0, 0, 1]^T \in \mathbb{R}^3\) and \(\mathcal{W}_w \in \mathbb{R}^3\) is continuous-time white Gaussian noise process. A geometric controller is employed to generate the thrust force (\(f_c \in \mathbb{R}\)) and the moment vector (\(M_c \in \mathbb{R}^{3}\)) necessary to maintain the quadrotor on its designated path \cite{Goodarzi2017}. The symbol \((\hat{.})\) represents the hat map. \(\Delta_1 \in \mathbb{R}^3\) and \(\Delta_2 \in \mathbb{R}^3\) account for the external forces and torques, which are not explicitly modeled.  \(\Delta_1\) and \(\Delta_2\) are considered fixed but unstructured.

In Equation \eqref{eq:vdot_w1}, \(D_f \in \mathbb{R}^3\) denotes the drag force, and the generic drag force is given by,
\begin{align}
    D_f = -\frac{1}{2}\rho_{air}C_d|v_r|v_r,
    \label{eq:dragf_w1}
\end{align}
where \(C_d \in \mathbb{R}^{3 \times 3}\) is the body drag coefficient matrix of the quadrotor UAV. The relative wind velocity in the body frame \(v_r \in \mathbb{R}^3\) is given by \(v_r = v_b - R^Tv_w\), where \(v_b \in \mathbb{R}^3\) is the vehicle velocity in the body frame. The velocity \(v_b\) can be expressed in terms of the inertial velocity \(v\) as, \(v_b = R^Tv\).


\paragraph{Discrete-Time Modeling of Quadrotor UAV}
The translational and attitude dynamics of  a quadrotor UAV, in discrete-time, subject to wind disturbances, are as follows \cite{LeeLeoCMDA07},
\begin{align}
h (J\Omega_k)^\wedge=& \mathcal{F}_k J_d-J_d \mathcal{F}_k^T,\label{eqn:eom0}\\
R_{k+1}=&R_k \mathcal{F}_k,\label{eq:Rdot_wd}\\
J \Omega_{k+1}=& \mathcal{F}_k^T J \Omega_k+hM_{e_k},\label{eq:Wdot_wd}\\
x_{k+1}=&x_k+hv_k,\label{eq:xdot_wd}\\
mv_{k+1}=&mv_k+hU_{e_k},\label{eq:vdot_wd}\\
v_{w_{k+1}} =& v_{w_{k}} + h\dot{v}_{w_k}, \label{eq:vwdot_wd}
\end{align}
where the subscript $k$ denotes the value of a variable at time $t_k=kh$, for $k\in\{1,\ldots, N\}$, with a fixed time step $h>0$. The matrix $J_d=\frac{1}{2}tr[{J}] I_{3\times 3}-J\in\mathbb{R}^{3\times 3}$ is the non-standard inertia matrix, and $\mathcal{F}_k\in\mathrm{SO}(3)$ is to represent the relative attitude change between two adjacent time steps. For each time step, first, Equation \eqref{eqn:eom0} is solved for $F_k$~\cite{LeeLeoCMDA07}, then \eqref{eq:Rdot_wd}-\eqref{eq:vwdot_wd} yield a discrete-time map $(R_k,\Omega_k,x_k,v_k, v_{w_{k}})\rightarrow(R_{k+1},\Omega_{k+1},x_{k+1},v_{k+1}, v_{w_{k+1}})$, which is repeated until the terminal time step $t_N=Nh$. 
Symbols \(U_{e_k} \in \mathbb{R}^3\) and \(M_{e_k} \in \mathbb{R}^3\) represent the resultant force resolved in the inertial-frame, and the resultant moment resolved in body-frame. \(U_{e_k}\) and \(M_{e_k}\) are expressed as , \(  U_{e_k} = mge_3 - f_cRe_3 + RD_f + \Delta_1\), and \(M_{e_k} = M_c + \Delta_2\).

\subsection{Measurements and Inputs}
Measurements \(x_{gps}\), \(\Omega_{imu}\), and \(a_{imu}\in \mathbb{R}^ 3\) are obtained from the low cost sensors, including the GPS and IMU. The measurement vector is \(Y=[x_{gps}, \Omega _{imu}, a_{imu}]\in \mathbb{R}^9\), which are available on low cost UAVs. Acceleration measurement \(a_{imu}\) is directly used in updating the wind velocity \(v_w\).

\subsection{Estimation}
Estimation of the full state vector \(X\), in the continuous and discrete-time domains, is performed using EKF and UKF. The estimation occurs is in two stages: the flow update and measurement update. During flow update, the posteriori state estimate (\(^+\)) of the previous time-step (\(k-1\)), \(\bar{X}_{k-1}^+\) and the posteriori error covariance matrix  \(P_{k-1}^+\) are propagated to obtain the priori estimates (\(^-\)), \(X^-_k \)  and \(P^-_k\), for the current time-step (\(k\)). The priori state estimates are then adjusted during the measurement update.

Wind dynamics are often modeled as a stochastic process in the literature, using methods such as Random Walk, Gauss-Markov or quasi-steady state assumption \cite{
Rhudy2019}. In this study, Random walk method is adopted, as the wind measurements are unknown and changing. 

 If the drag force \(D_{f_k}\) at the current time step is known, priori estimate of the wind velocity \(v_{w_k}^-\) can be obtained using Equation \eqref{eq:dragf_w1}. Though the drag force \(D_f\) is unknown directly, we can approximate it using the acceleration in the absence of the drag force  \(a_{ideal}\in \mathbb{R}^3\), and the acceleration measurement \(a_{imu}\) from the IMU. The accelerations \(a_{ideal}\) and \(a_{imu}\) can be expressed as follows, 
\begin{align}
     a_{imu} &= v_b \times \Omega - \frac{f_ce_3}{m} + \frac{D_f}{m} + R^T\frac{\Delta_1}{m},\\
     a_{ideal} &= v_b \times \Omega - \frac{f_ce_3}{m} + R^T\frac{\Delta_1}{m}.
 \end{align}
We assume that the accelerometer measures the non-gravitational acceleration in the body frame. Therefore, the acceleration relative to the world frame with gravity is neglected. Thus, the drag force \(D_{f_k}\) can be approximated as follows, 
\begin{gather}
     D_{f_k} = m(a_{IMU_k}-a_{ideal_k}),
     \label{eq:Dragf_WI1}
\end{gather}
 By calculating \(D_{f_k}\), a priori wind velocity \(v_{w_k}^-\) can be estimated using \eqref{eq:dragf_w1}. 
 The states \(X^+_{k-1}\) now containing the updated \(v_{w_{k}}^-\), are propagated through Equations \eqref{eq:xdot_w1} to \eqref{eq:vwdot_wd}, in the prediction stage. At the measurement update stage, \(x_{gps}\), \(\Omega_{imu}\), and \(a_{imu}\) are used to correct the predictions \(X^-_k\), and obtain the final estimation \(X^+_k\). 

\paragraph{EKF}
The EKF algorithm relies on a state-dependent linearization of the nonlinear quadrotor system. The system is linearized using Jacobian-based perturbations, \(\delta \dot{X} = [\delta \dot{x}, \delta \dot{v}, \delta \dot{\Omega}, \delta \dot{R}, \delta \dot{v}_w]\). In geometric control, the rotation matrix \(R\) is not directly perturbed, as it should remain on \(SO(3)\). Instead, we express perturbations with a minimal vector \(\eta \in \mathbb{R}^3\) through the exponential map \cite{Gamagedara2019}. Using first-order approximation, \(\eta\) is defined so that it satisfies,
\begin{gather}
       \delta R = \left. \frac{d}{d\epsilon} \right|_{\epsilon=0} R \exp(\epsilon \hat{\eta}) = R \hat{\eta},
   \label{eq:dR}
\end{gather}
where \(\epsilon \in \mathbb{R} \) is a smaller scalar parameter controlling perturbation magnitude. Therefore, the perturbed state vector \(\delta \dot{X} = [\delta \dot{x}, \delta \dot{v}, \delta \dot{\Omega}, \dot{\eta}, \delta \dot{v_w}]\in \mathbb{R}^{15}\) is used in the Jacobian linearization. 

 Focusing on the deterministic system dynamics, perturbed form of Equation \eqref{eq:xdot_w1} can be expressed as follows,
\begin{align}
    \delta \dot{x} &= \delta v. \label{eq:dxdot_w1}
    \end{align}
    The perturbed form of Equation \eqref{eq:vdot_w1} can be expressed as follows,
    \begin{align}
     \delta \dot{v} &= \delta \big(- \frac{f_cRe_3}{m}\big) +  \delta \big(\frac{RD_f}{m}\big) ,\notag
     \end{align}
where, 
\begin{align*}
    RD_f &= -\frac{1}{2}\rho_{air}C_dR|R^Tv-R^Tv_w|(R^Tv-R^Tv_w),\\
     &= -\frac{1}{2}\rho_{air}C_d|v-v_w|(v-v_w).
\end{align*}
 The system was linearized around \(v_w = 0\), as the nominal wind is assumed unknown and also to simplify the analytic Jacobians. Thus, \(\dot{v}\) can be expressed as,
\begin{align}
    \delta \dot{v} &=   -\frac{1}{2m}\rho_{air}C_d(|v|I_{3\times 3} + \frac{vv^T}{|v|})\delta v + \frac{f_c}{m} R(e_3)^{\wedge}\eta ,  \label{eq:dvdot_w1}
    \end{align}
    where \(\frac{\partial|v|}{\partial v}=\frac{v^T}{|v|}\), and \( \delta (Re_3)= R\hat{\eta}e_3 =  R(e_3)^{\wedge}\eta\). As \(\eta\) is already a perturbation itself.
    Therefore, using Equation \eqref{eq:Rdot_w1} and \eqref{eq:dR}, \(\dot{\eta}\) can be defined. From Equation \eqref{eq:Rdot_w1}, 
    \begin{align}
        \delta \dot{R} = R\hat{\eta}\hat{\Omega} + R(\delta \hat{\Omega}). \notag
    \end{align}
    From the relationship in Equation \eqref{eq:dR},
    \begin{align}
        \delta \dot{R} &= \dot{R}\hat{\eta}+R\hat{\dot{\eta}}. \notag
    \end{align}
    By equating both expressions for \(\delta \dot{R}\), and substituting from Equation \eqref{eq:Rdot_w1}
    \begin{align}
        \dot{R}\hat{\eta}+R\hat{\dot{\eta}} &= R\hat{\eta}\hat{\Omega} + R(\delta \hat{\Omega}), \notag \\
        R\hat{\dot{\eta}} &=  R(\delta \hat{\Omega}) + R\hat{\eta}\hat{\Omega} - R\hat{\Omega}\hat{\eta}, \notag \\
        \hat{\dot{\eta}} &= (\delta \hat{\Omega}) + \hat{\eta}\hat{\Omega} - \hat{\Omega}\hat{\eta}. \notag
    \end{align}
    Given that \(-\hat{\eta}\hat{\Omega} + \hat{\Omega}\hat{\eta} = \widehat{\Omega \times \eta}\) ,
\begin{align}
    \hat{\dot{\eta}} & = \delta \hat{\Omega} - \widehat{\Omega \times \eta}, \notag \\
    {\dot{\eta}} & = \delta {\Omega} - {\Omega \times \eta}, \notag \\
    \dot{\eta} &= -\hat{\Omega}\eta + \delta \Omega,  \label{eq:dRdot_w1}
    \end{align}
where, \(\Omega \times \eta = \hat{\Omega}\eta\). The perturbed state \(\delta \dot{\Omega}\) is obtained by linearizing the rotational dynamics given in  Equation \eqref{eq:Wdot_w1} and neglecting the first-order variations of the control/disturbance torque yields, as follows, 
    \begin{align}
    \delta \dot{\Omega} &= \delta  (J^{-1}(-\Omega \times J\Omega + M_{c} )),\notag\\
    \delta \dot{\Omega} &= J^{-1}\delta (-\Omega\times J\Omega) . \notag 
    \end{align}
    As \(-\Omega\times J\Omega = -\hat{\Omega}J\Omega = (J\Omega)^\wedge\Omega\),
    \begin{align}
    \delta \dot{\Omega} &= J^{-1}(-\hat{\Omega}J\delta\Omega + (J\Omega)^\wedge\delta\Omega) +J^{-1}\delta \mathcal{W}_{\Omega}, \notag\\
        \delta \dot{\Omega} &= J^{-1}((J\Omega)^\wedge - \hat{\Omega}J)\delta \Omega + J^{-1}\delta \mathcal{W}_{\Omega}.  \label{eq:dwdot_w1}
\end{align}
The perturbed wind dynamics can be expressed as follows,
    \begin{align}
    \delta \dot{v}_w &= 0 .  \label{eq:dvwdot_w1}
\end{align}
The second-order perturbations have been neglected in this study. 

The stochastic components of the system dynamics are \(\mathcal{W}=[\mathcal{W}_v, \mathcal{W}_\Omega, \mathcal{W}_w]^T \in \mathbb{R}^9\). which affect only \(\dot{v}\), \(\dot{\Omega}\), and \(\dot{v}_w\).  In the current time-step, Equations \eqref{eq:dxdot_w1} to \eqref{eq:dvwdot_w1}, can be expressed in the linearized form \(\delta \dot{X}= {F}_{t_{k-1}}\delta X+ {G}_{t_{k-1}}\mathcal{W}\), where the state transition matrix \({F}_{t_{k-1}} \in \mathbb{R}^{15 \times 15}\) and process noise transition matrix \({G}_{t_{k-1}}\in \mathbb{R}^{15 \times 9}\) are as follows, 
 
 \begin{align}
        {F}_{t_{k-1}} &=
    \begin{bmatrix}
        0_{3} & I_{3} & 0_{3} & 0_{3} & 0_{3} 
        \\
        0_{3} & F_{1} & F_{2} & 0_{3} & 0_{3}  
\\
        0_{3} & 0_{3} & F_{3} & I_{3} & 0_{3}   
\\
        0_{3} & 0_{3} & 0_{3} & F_{4} & 0_{3} 
\\
        0_{3} & 0_{3} & 0_{3} & 0_{3} & 0_{3} 
\\
    \end{bmatrix},\\
    {G}_{t_{k-1}} &= 
    \begin{bmatrix}
        0_{3} & 0_{3}  & 0_{3}\\
        \frac{I_{3}}{m} & 0_{3} & 0_{3} \\
        0_{3} & 0_{3} & 0_{3}  \\
        0_{3} & J^{-1} & 0_{3} \\
         0_{3} & 0_{3} & I_{3}
    \end{bmatrix}, \label{eq:G_WI1}
\end{align}
where,
\begin{align*}
    F_{1} =&   -\frac{1}{2m}\rho_{air}C_d(|v_{k-1}|I_{3\times 3} + \frac{v_{k-1}v_{k-1}^T}{|v_{{k-1}|}}),\\
    F_{2} =&  (f_{c_{k-1}}/m) R_{k-1}(e_3)^{\wedge},\\
    F_{3} =& -\hat{\Omega}_{k-1},\\
    F_{4} =& J^{-1}((J\Omega_{k-1})^\wedge - \hat{\Omega}_{k-1}J).
\end{align*}
The general linearized discrete form is \( X_k^- = {F}_{k-1}X_{k-1}^+ + {G}_{k-1}\mathcal{W}_{k-1}\), where \(\mathcal{W}_{k-1}\in \mathbb{R}^{9}\) represents the zero-mean Gaussian white process noise. The matrices \({F}_{t_{k-1}}\) and \({G}_{t_{k-1}}\) are transformed into their respective discrete forms \({F}_{k-1}^c\) and \({G}^c_{k-1}\) using the following relationship,
\begin{align}
    {{F}}_{t_{k-1}} &\equiv \left. \frac{\partial{f}}{\partial{X}} \right|_{\dot{X}_{(t)},\mathcal{U}_{(t)}},\quad
    {{G}}_{t_{k-1}} \equiv \left. \frac{\partial{\mathcal{G}_{(t)}}}{\partial{X}} \right|_{\dot{X}_{(t)},\mathcal{U}_{(t)}}, \label{eq:FG_c}\\  
    {{F}}_{k-1}^c &= I_{n \times n} + h {{F}}_{t_{k-1}}\Psi, \quad {{G}}_{k-1}^c = h\Psi {{G}}_{t_{k-1}}, \label{eq:FGc1}
\end{align}
where, \(\Psi = I_{n \times n} + \frac{h}{2}{F}_{t_{k-1}}\).

As for the quadrotor dynamics in discrete-time, the perturbed states at current time step \(\delta X_k\) can be expressed as follows,
\begin{align}
    \delta {x}_{k+1} &= \delta x_k +h\delta v_k, \label{eq:dx_w1}\\
    \delta {v}_{k+1} &=   \delta v_k + h(\delta\dot{v}_k),  \label{eq:dv_w1}
    \end{align}
    where, \(\delta \dot{v}_k\) is as in Equation \eqref{eq:dvdot_w1}. To obtain the \(\eta_{k+1}\), the relationship in Equation \eqref{eq:dR} and a first-order small angle approximation,  \(\mathcal{F}_{k+1} \approx I+h\hat{\Omega} \) is used. A small increase in attitude \(\delta R_{k+1}\) can be expressed as follows using Equation \eqref{eq:Rdot_wd}, 
    \begin{equation}
        \delta R_{k+1} = R_k\hat{\eta}_k + R_kh\delta\hat{\Omega}_k. \notag
    \end{equation}
    Similarly, using Equation \eqref{eq:dR}, 
    \begin{align}
    \delta R_{k+1} = R_{k+1}\hat{\eta}_{k+1}\\
    = (R_k\mathcal{F}_k)\hat{\eta}_{k+1} \\
        \delta R_{k+1} = R_k \hat{\eta}_{k+1}. \notag
    \end{align}
    Thus, equating the expressions for  \(\delta R_{k+1}\), \(\eta_{k+1}\) can be defined as follows, 
    \begin{align}
  R_k \hat{\eta}_{k+1} &= R_k\hat{\eta}_k + R_kh\delta\hat{\Omega}_k  ,\notag\\
    \hat{\eta}_{k+1} &= \hat{\eta}_k + h\delta \hat{\Omega}_k, \notag \\
    {\eta}_{k+1} & \approx  \eta_k +h{\delta\Omega_k},  \label{eq:deta_w1}
    \end{align}
    where second-order perturbations are neglected. Perturbation of \(\Omega_{k+1}\) can be expressed as follows,
    \begin{align}
    \delta {\Omega}_{k+1} &= \delta \Omega_k +hJ^{-1}((J\Omega_k)^\wedge - \hat{\Omega}_kJ)\delta \Omega_k,  \label{eq:dw_w1}
    \end{align}
    where,  \(\hat{\Omega}J\Omega = -(J\Omega)^\wedge\Omega\). Finally the perturbation of wind velocity \(\delta v_{w_{k+1}}\) can be defined as,
    \begin{align}
    \delta {v}_{w_{k+1}} &= {v}_{w_{k}}.  \label{eq:dvw_w1}
\end{align}
Since the discrete-time formulation of the perturbed states are already in the discrete form  the state transition matrix \({F}^d_{k-1}\) and process noise transition matrix \({G}^d_{k-1}\) can be directly derived from Equations \eqref{eq:dx_w1} to \eqref{eq:dvw_w1}, as follows,
 \begin{align}
        {F}^d_{{k-1}} &=
    \begin{bmatrix}
        I_{3} & hI_{3} & 0_{3} & 0_{3} & 0_{3} 
        \\
        0_{3} & F_{1} & F_{2} & 0_{3} & 0_{3}  
\\
        0_{3} & 0_{3} & I_{3} & hI_{3} & 0_{3}   
\\
        0_{3} & 0_{3} & 0_{3} & F_{3} & 0_{3} 
\\
        0_{3} & 0_{3} & 0_{3} & 0_{3} & I_{3} 
\\
    \end{bmatrix},\\
    {G}^d_{{k-1}} &= 
    \begin{bmatrix}
        0_{3} & 0_{3} & 0_{3}  \\
        \frac{hI_{3}}{m} & 0_{3} & 0_{3} \\
        0_{3} & 0_{3} & 0_{3}  \\
        0_{3} & hJ^{-1} & 0_{3} \\
         0_{3} & 0_{3} & hI_{3}
    \end{bmatrix}, \label{eq:Gd_WI1}
\end{align}
where,
\begin{align*}
    F_{1} =&   I_3 -h\bigg(\frac{1}{2m}\rho_{air}C_d(|v_{k-1}|I_{3\times 3} + \frac{v_{k-1}v_{k-1}^T}{|v_{{k-1}|}})\bigg),\\
    F_{2} =&  h(f_{c_{k-1}}/m) R_{k-1}(e_3)^{\wedge},\\
      F_{3} =& I_3 + hJ^{-1}((J\Omega_{k-1})^\wedge - \hat{\Omega}_{k-1}J).
\end{align*}

The continuous-discrete form of the EKF is adapted \cite{Crassidis2004} in this study. Thus, the priori estimate of the states \(X^-_k\) for the continuous-time, is obtained by propagating the previous state estimate \(X^+_{k-1}\), through Equations \eqref{eq:xdot_w1} to \eqref{eq:vwdot_w1} using Runge-Kutta fourth-order integration, while for the discrete-time, through Equations \eqref{eq:dxdot_w1} to \eqref{eq:dvwdot_w1}. Propagation through nonlinear dynamics, improves the accuracy of the priori estimate. The rest of the EKF algorithm applies the same for both continuous and discrete formulations. The priori estimate of the error covariance matrix \(P^-_k\) is follows, 
\begin{gather}
     P_k^- = {F}_{k-1}P_{k-1}{F}_{k-1}^T + {G}_{k-1}{Q}_k{G}_{k-1}^T,
    \label{eq:P-kj}
\end{gather}
where, \(\mathcal{Q}_k=\mathbb{E}[\mathcal{W}_k\mathcal{W}_k^T]\in\mathbb{R}^{9 \times 9}\) is the process noise covariance matrix.

During the measurement update stage, the Kalman gain \(K_k \in \mathbb{R}^{15 \times 9} \) is calculated as follows,
\begin{align}
    K_k = P_k^-H_k^T[H_k P_k^-H_k^T+\mathcal{R}_k]^{-1},
\end{align}
where, \(H_{k} = \left.\frac{\partial{h}}{\partial{X}} \right|_{\bar{X}_k^-} \in \mathbb{R}^{9 \times 15}\), is the Jacobian of the measurement function, and \(\mathcal{R}_k=\mathbb{E}[\mathcal{V}_k\mathcal{V}_k^T]\in\mathbb{R}^{9 \times 9}\). Linearized form of the general measurement equation is expressed as \(Y_k = H_kX_k^- + \mathcal{V}_k\),
where, \(\mathcal{V}_{k}\in \mathbb{R}^{9}\) represents the zero-mean Gaussian white measurement noises. 
The measurement update is carried out in modular form where the priori estimates \(R^-_k\) and  \(\Omega^-_k\) are updated using angular velocity measurements \(\Omega_{imu}\). The position \(x^-_k\) and velocity \(v^-_k\) priori estimates are updated using measurements from GPS, \(x_{gps}\). Accelerometer measurement \(a_{imu}\) is only used in updating the wind velocity \(v_{w_k}^-\). The true measurements from the sensors \(Z_k =[x_{gps}, \Omega_{imu}, a_{imu}]\in \mathbb{R}^9\), are used in updating the posteriori states \(X_k^+\), as follows,
\begin{align}
    \bar{X}_k^+=\bar{X}_k^-+K_k[Z_k-H_k\bar{X}_k^-].
\end{align}
Finally, the posteriori error covariance matrix \(P^+_k\) is updated as follows,
\begin{gather}
    P_k^+=[I-K_kH_k]P^-_k[I-K_kH_k]^T+K_k\mathcal{R}_kK_k^T.
    \label{eq:P+kj}
\end{gather}
Equation \eqref{eq:P+kj}, Josheph's form for error covariance is used to ensure that the error covariance matrix \(P^+_k\) remains positive definite and symmetric \cite{Gamagedara2019, Crassidis2004}.

\paragraph{UKF}
The UKF is a variation of the Kalman filter designed to address the nonlinearities of a discrete dynamic system. In UKF, a set of strategically chosen points around the mean of the states, known as sigma points, plays a major role. The sigma points are propagated through the nonlinear system functions. Thus, allowing to capture the transformation of the mean and covariance directly. The full state vector \(X=[x, v, \Omega, R, v_w]\in \mathbb{R}^{21}\) is used in the UKF algorithm. As the measurement noise is additive, a non-augmented UKF algorithm was used in this study \cite{Haykin2001}.

During measurement update stage, the posteriori state estimates from the previous time step are used to generated the sigma points \((\mathcal{X}^i_{k-1} \in \mathbb{R}^{n \times 2n+1}, \quad i=0,...,2n)\), such that \(\mathcal{X}_{k-1} = \bigg[ X_{k-1}^+, \quad X_{k-1}^+ \pm \sqrt{(n + \lambda)P_{k-1}^+} \bigg]\). Here, the scaling parameter \(\lambda\in \mathbb{R}\) is calculated such that \(\lambda=\alpha^2(n+\kappa)-n\), where \(\alpha=1\) ensures sufficient spread of the sigma points, and \(\kappa \in \mathbb{R}\) determines the secondary scaling. In this study \(\alpha=1\), to ensure the sigma points are spread sufficiently, \(\kappa =0\) is adopted since additional scaling is unnecessary. A larger \(\lambda\) spreads the points more, captures more nonlinearity, but at a higher risk of the system becoming unstable. There are \(n=21\) states, thus a total of \(43\) sigma points are generated. The generated sigma points \(\mathcal{X}^i_{k-1}\), are then propagated through the nonlinear dynamic equations of the quadrotor UAV, Equations \eqref{eq:xdot_w1} to \eqref{eq:vwdot_w1} for the continuous form(using Runge-Kutta fourth-order integration), and Equations \eqref{eq:dxdot_w1} to \eqref{eq:dvwdot_w1} for the discrete form. For the rest of the UKF algorithm, both continuous and discrete forms follow the same procedure. Propagated sigma points are then used to calculated the priori mean estimate of the full state vector \(\bar{X}^-_k \in \mathbb{R}^{21}\), and priori estimate of the error covariance \(P^-_k\) as follows,
\begin{align}
    \bar{X}_k^- &= \sum _{i=0}^{2n} W_i^{mean} \mathcal{X}_k^i, \\
    P_k^- &= \sum_{i=0}^{2n} W_i^{cov}(\mathcal{X}_k^i - \bar{X}_k^-)(\mathcal{X}_k^i - \bar{X}_k^-)^T + \mathcal{Q}_k,
\end{align}
where, \(W_{0}^{mean} = \frac{\lambda}{n + \lambda}\), \(W_{0}^{cov} = \frac{\lambda}{n + \lambda}(1 - \alpha ^2 + \beta)\), and \(W_{i}^{mean} = W_{i}^{cov} = \frac{1}{2(n + \lambda )}\in \mathbb{R}, \quad i = 1, 2, . . ., n\).  The covariance weight for the sigma points are denoted by \(\beta\) . For Gaussian distributions, \(\beta\) is considered to be 2. 

During the measurement update stage, the priori estimates \(\bar{X}^-_k\) and \(P^-_k\) are updated. For that, first the mean predicted measurement \(\bar{Y}_k^-\in \mathbb{R}^9\) should be obtained. Hence, the sigma points \(\mathcal{X}^i_k\), are transformed using the general measurement equation, \(Y_k = h(X_k, \mathcal{U}_k, \mathcal{V}_k, k)\). The transformed sigma points \(\mathcal{Y}^i_k \in \mathbb{R}^9\) are then used as follows,
\begin{align}
     \bar{Y}_k = \sum_{i=0} ^{2n} W_{i}^{mean} \mathcal{Y}_k^i. 
\end{align}
Subsequently, Kalman gain \(K_k \in \mathbb{R}^{21 \times 6}\) is calculated as follows,
\begin{align}
    P_{yy_k} &= \sum_{i=0}^{2n} W_i^{cov} (\mathcal{Y}_k^i - \bar{Y}_k)(\mathcal{Y}_k^i - \bar{Y}_k)^T +\mathcal{R}_k,\\
    P_{xy_k} &= \sum _{i=0}^{2n} W_i^{cov}(\mathcal{X}_k^i - \bar{X}_k^-)(\mathcal{Y}_k^i - \bar{Y}_k )^T,\\
    K_k &= P_{xy_k} P_{yy_k}^{-1}.
\end{align}
Finally, posteriori state estimate \(\bar{X}^+_k\) and the error covariance \(P^+_k\) are calculated as follows,
\begin{align}
    \bar{X}_k^+ = \bar{X}_k^{-} + K_k(Z_k - \bar{Y}_k),\\
    P_k^+ = P_k^- + K_kP_{yy_k}K_k^T.
\end{align}

\subsection{Implementation Settings}

The parameters of the quadrotor used in the simulation are included in Table \ref{tab:quadrotor_parameters}.  The quadrotor UAV is equipped with an ICM-45686 (6-axis IMU) \cite{icm45686} with gyroscope accuracy \(3.8\;mdps/\sqrt{Hz}\) and accelerometer accuracy \(70\;\mu g/\sqrt{Hz}\), as well as an M10 GPS module \cite{m10gps} with accuracy \(2.0\;CEP\). The corresponding standard deviations are summarized in Table \ref{tab:quadrotor_parameters}. However,  for the present small-trajectory experiments we used the IMU and accelerometer values from Table \ref{tab:quadrotor_parameters} directly, the nominal GPS variance was replaced with \(\sigma_{gps}=0.1 m\) to reflect the higher positioning quality during the simulated flights.
\begin{table}
\centering
    \caption{Parameters of the quadrotor model}
    \label{tab:quadrotor_parameters}
\begin{tabular}{|p{4.8cm}|p{1.6cm}|c|}
    \hline
    \textbf{Symbol: Description} & \textbf{Value} & \textbf{Unit} \\ \hline
    
    $I_{xx}$: MOI about body frame's x-axis & $0.02$ & $\mathrm{kg{\cdot}m^2}$ \\ \hline
    
    $I_{yy}$: MOI about body frame's y-axis & $0.02$ & $\mathrm{kg{\cdot}m^2}$ \\ \hline
    
    $I_{zz}$: MOI about body frame's z-axis & $0.04$ & $\mathrm{kg{\cdot}m^2}$ \\ \hline
    
    $d$: Distance from rotor arm to the quadrotor's center of gravity & $0.169$ & $m$ \\ \hline
    
    $m$: Quadrotor mass & $2.0$ & $kg$ \\ \hline

    $C_d$: Body drag force coefficient & diag([ 1.1, 1.1, 0.55]) &   \\ \hline

    $\sigma_{gps}$: GPS standard deviation & $0.100$ & $m$  \\ \hline

    $\sigma_{gyr}$: Gyroscope standard deviation & $0.009$ & $ rad/s$ \\ \hline

    $\sigma_{acc}$: Accelerometer standard deviation & $0.097$ & $ ms^{-2}$ \\ \hline

    $f_{gps}$: GPS frequency & $5$ & $Hz$  \\ \hline

    $f_{imu}$: IMU frequency & $400$ & $ Hz$ \\ \hline

    $g$: Gravitational acceleration & $9.81$ & $ms^{-2}$ \\ \hline
    \end{tabular}
\end{table}
The system was introduced with fixed disturbances \(\Delta_1\) and \(\Delta_2\) as process noise and non-additive, zero-mean, Gaussian noise as measurement noise,  which affects the measurements from the GPS, gyroscope, and accelerometer. Equations \eqref{eq:vdot_w1} and \eqref{eq:Wdot_w1} include fixed disturbances, \(\Delta _1 = [    0.5,  0.8, -1.0], \: \: \Delta _2 = [0.2, 1.0, -0.1]\).

\subsubsection{Numerical Simulation}
The performance of EKF and UKF estimations was tested using numerical simulations using MATLAB. The quadrotor UAV was tested while hovering,as well as while following a Lissajous trajectory. Lissajous trajectory was used to increase the system nonlinearity, thus gauging the performance of the estimators in a highly nonlinear scenario. During hovering, the quadrotor started at an initial point \( x_i = [0, 0, 0]^T\) and traveled to  \( x_d = [1, 0, -1]^T\), and hovered at the point. The desired trajectory of the Lissajous curve is \(x_{d(t)} = [ sin(t ), sin(2t),  -1 + 0.2cos(2t)]^T\).
The quadrotor UAV was disturbed with a constant wind field of \( v_{wt} = [4, 5, 0]^T\). To gauge the estimation accuracy in a more complex wind field, it was also disturbed with a sinusoidal wind field, which is \(v_{wt} =[5\sin(2\pi f_1 t), 4\sin(2\pi f_2 t), 4\sin(2\pi f_2 t)]^T \), where, \(f_1 = 1/15 \: Hz\) and \(f_2 = 2/15 \: Hz\). 

\subsubsection{Experimental Setup}

\begin{figure}[h]
    \centering
    \includegraphics[width=1\linewidth]{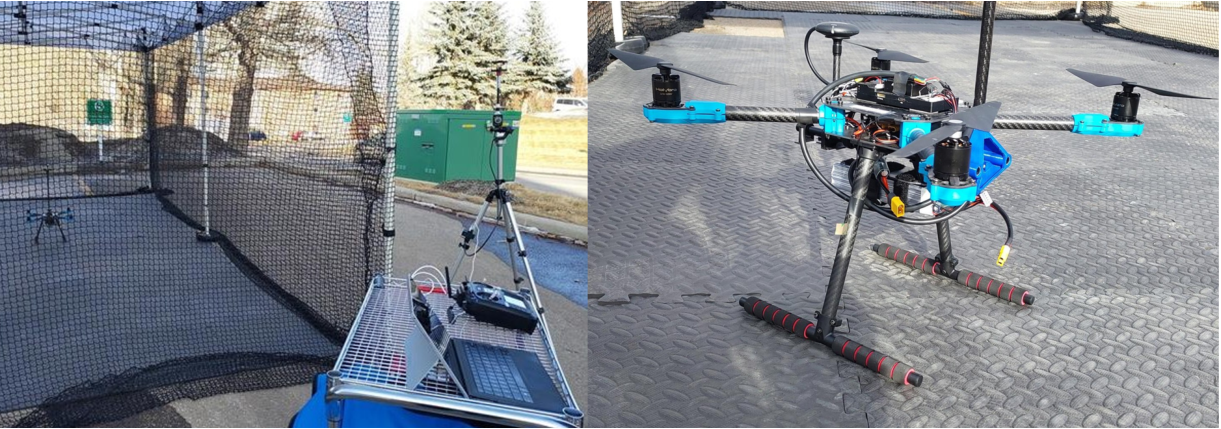}
    \caption{Experimental Setup for Operating the Holybro X500 V2 quadrotor UAV}
    \label{fig:Experiment_setup}
\end{figure}

The experiments were conducted on February \(3^{rd}\), 2026 at the Mechanical Engineering Building of Schulich School of Engineering, University of Calgary. A Holybro X500 V2 quadrotor UAV, equipped with a M10 GPS and ICM-45686 was used in wind estimation. For validation, two LI-550 TriSonica Mini sonic anemometers were used, one mounted onto the quadrotor UAV, and the other as a ground-based measurement. The ground-based anemometer was placed 1.5 m above the ground. The quadrotor UAV was tested while hovering as well as whole moving, inside 5 m radius of the ground-based anemometer, and the arrangement is indicated in Figure \ref{fig:Experiment_setup}.

\section{Results and Discussion}

\subsection{Numerical Simulation}

In the numerical simulations, both additive and non-additive noises were simulated to account for measurement noise, as well as for fixed-disturbances. Thus, wind velocity estimation using quadrotor UAV, is carried out in two stages. The first stage is for fixed-disturbance identification. The quadrotor UAV hovers, fly through a straight-line trajectory and through a Lissajous  curve, in the absence of wind disturbances, during which the fixed disturbances are estimated using EKF and UKF. The estimates from each trajectory are then averaged to obtain an final estimate for the fixed disturbances, which are shown in Table \ref{tab:bias_estimate}. Continuous-time and discrete-time formulations are represented by -C and -D respectively.
\begin{table}[htb!]
    \centering
    \caption{Estimations for bias of a Quadrotor UAV}
    \label{tab:bias_estimate}
    \begin{tabular}{|c|c|c|c|c|c|c|}
    \hline
    \textbf{Type} & \multicolumn{3}{|c|}{\(\Delta_1\)} & \multicolumn{3}{|c|}{\textbf{\(\Delta_2\)}} \\ \hline
    EKF-C & 0.516& 0.776& -0.992& 0.200& 1.000& -0.100\\ \hline
    EKF-D & 0.516& 0.778& -0.992& 0.200& 1.000& -0.100\\ \hline
    UKF-C & 0.512& 0.802& -0.978& 0.200& 1.000& -0.100\\ \hline
    UKF-D & 0.504& 0.800& -0.986& 0.200& 1.000& -0.100\\ \hline
    True & 0.500& 0.800& -1.000& 0.200& 1.000& -0.100\\ \hline
              \end{tabular}
\end{table}

In this work, the drag moment is neglected, considering its negligible effects, compared to the drag force. The measurement noise corrupting the measurements is assumed to be as given in \ref{tab:quadrotor_parameters}.  
Table \ref{tab:quadrotor_parameters} reports measured and manufacturer noise figures across multiple environments. Though,  for the present small-trajectory experiments we used the IMU and accelerometer values from Table \ref{tab:quadrotor_parameters} directly, the nominal GPS variance was replaced with \(\sigma=0.1 m\) to reflect the higher positioning quality during the simulated flights. The quadrotor UAV continuous and dynamic models are simulated at 400 \(Hz\). The IMU delivers measurements at the same rate, while the GPS update occurs at 5 \(Hz\).

During the second stage, quadrotor UAV estimates the wind relying on the corresponding estimates from stage one. To assess the performance of quadrotor UAV with EKF and UKF algorithms, the wind velocity estimation is carried out in three different cases.
\begin{itemize}
    \item Case 1: Quadrotor following a Lissajous trajectory under constant wind disturbances, 
    \item Case 2: Quadrotor in hovering under sinusoidal wind disturbances, and
    \item Case 3: Quadrotor following a Lissajous trajectory under sinusoidal wind disturbances. 
\end{itemize}

\begin{figure}
    \centering
    \includegraphics[width=1\linewidth]{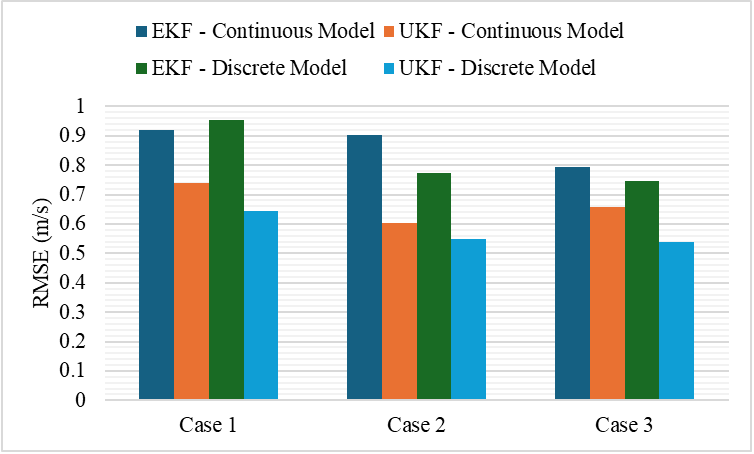}
    \caption{Performance of EKF and UKF when propagated using the continuous model and the discrete model of quadrotor UAV for the three testing scenarios}
    \label{fig:NS_results}
\end{figure}

We used sensor-based values for \(\mathcal{R}\) as listed in Table \ref{tab:quadrotor_parameters}, and adjustments to optimize estimation performance, while the process and measurement noise covariances were defined as (\(\mathcal{Q}_{EKF} \in \mathbb{R}^{6 \times 6}\) and \(\mathcal{Q}_{UKF} \in \mathbb{R}^{21 \times 21}\)), (\(\mathcal{R}_{EKF} \in \mathbb{R}^{9 \times 9} \), and \(\mathcal{R}_{UKF} \in \mathbb{R}^{6 \times 6}\)). 
The initial estimates for wind velocity \((v_w)\) contained high-frequency noise. Thus a second-order Butterworth filter is used as a postprocessing step. The corresponding results, evaluated via RMSE and standard deviation, are presented in the Figure \ref{fig:NS_results}.

The results indicate that the discrete quadrotor modeled estimated using UKF indicate higher accuracy, while the EKF implemented on the continuous model shows the worst performance. Additionally, across both continuous- and discrete-time formulations, the UKF tends to consistently outperform the EKF. The case 3, is the most nonlinear scenario, as it involves a quadrotor following a nonlinear path under nonlinear wind disturbances. As a nonlinear estimator, UKF better captures higher-order nonlinearities, which give an advantage when performing aggressive maneuvers under nonlinear conditions. In case 01(Lissajous trajectory under constant wind), overall wind velocity estimation accuracy is low, likely because the near-constant wind produces little excitation and is therefore harder to identify. The continuous model deals with integration residuals due to integrations, whereas discrete model avoids them, improving estimator consistency.

\subsection{Experimental Results}

\begin{figure}[h]
    \centering
    \includegraphics[width=1\linewidth]{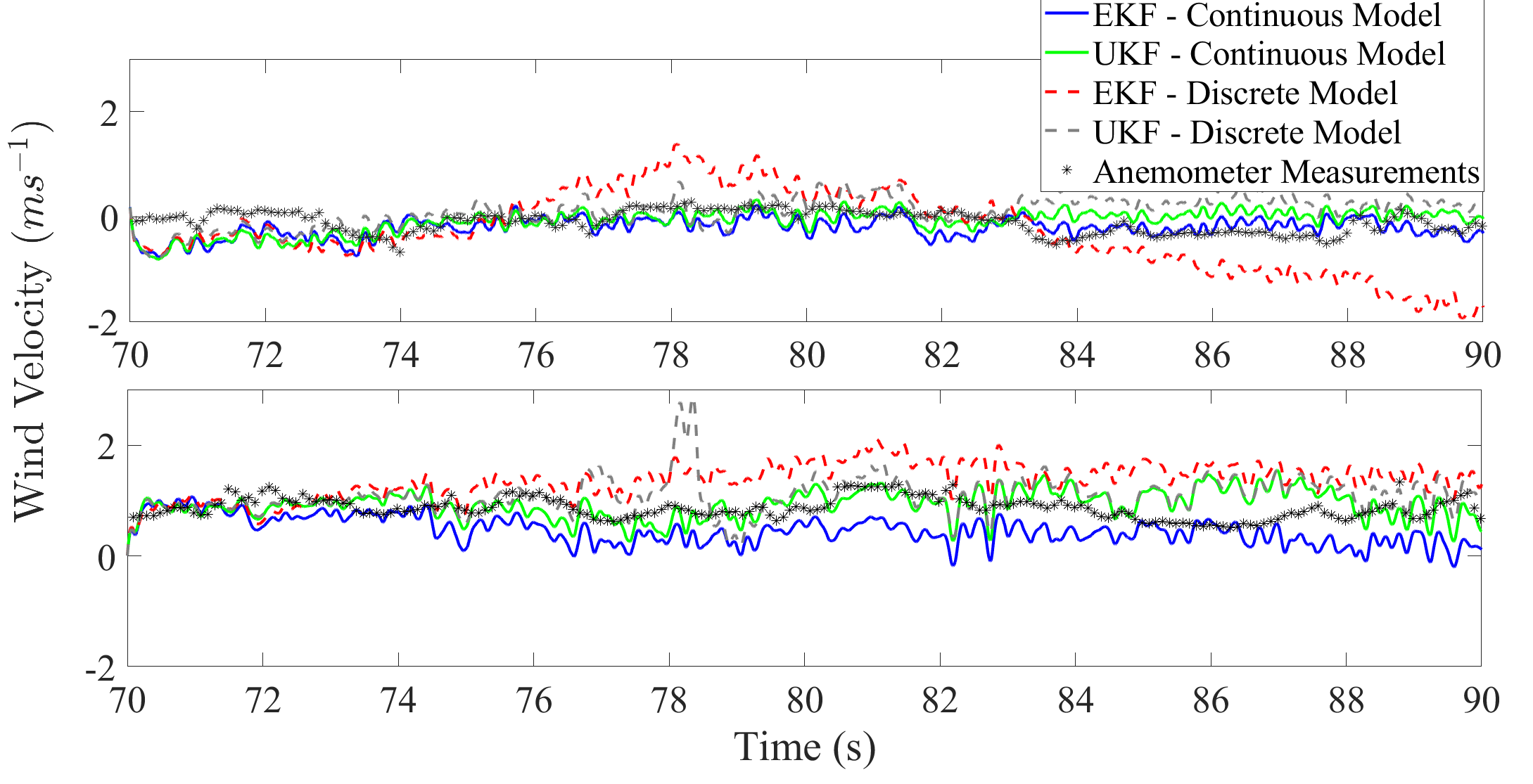}
    \caption{Performance of the 4 algorithms compared to the measurements from the ground-based anemometers}
    \label{fig:Experiment_results}
\end{figure}

The Holybro X500 V2 quadrotor UAV was manually operated with continuous small translational motions. The quadrotor was in motion for 10 minutes, and the recorded sensor measurements were processed sequentially the continuous EKF and UKF algorithms, which rely of continuous model and discrete model. Figure \ref{fig:Experiment_results} indicates the sample results for wind velocity estimation using 4 algorithms.

The performance of the each algorithm were evaluated by calculating the Root Mean Squared Error (RMSE). Evaluation of the algorithms for the complete duration is indicated in Table \ref{tab:Experiment_results}.

\begin{table}[h]
    \centering
        \caption{2D Wind Velocity Estimation using measurements from quadrotor UAV and comparing with Ground-based Anemometer}
    \label{tab:Experiment_results}
    \begin{tabular}{|c|c|c|}\hline
      \textbf{Estimator}   & \textbf{\(v_w\) in North} & \textbf{\(v_w\) in East}\\ \hline
       \textbf{EKF - Continuous}  & 0.6031 (\(\pm\) 0.5293) &  1.0335 (\(\pm\) 0.5387) \\ 
       \textbf{UKF - Continuous}  & 0.4715 (\(\pm\) 0.4534) & 1.0412 (\(\pm\) 0.7964) \\ 
       \textbf{EKF - Discrete}  & 2.0002 (\(\pm\) 1.0035) & 1.4663 (\(\pm\) 1.0558) \\ 
       \textbf{UKF - Discrete}  & 0.7631 (\(\pm\) 0.6145) & 0.7946 (\(\pm\) 0.6964) \\ \hline
    \end{tabular}
\end{table}
In contrast to the results from the numerical simulation, the wind velocity estimation using continuous model of the quadrotor indicate higher accuracy. Overall, UKF estimator on both model types show similar performance. Discrete model is more sensitive to noise and drift of the sensors. As the real sensors are suspetible to noise and drift it might cause the discrete model to shift away from the true dynamics. EKF applied on the discrete model of the quadrotor UAV has the least accuracy. As the EKF is based upon a linearized system model at each step, the nonlinearities of the system are often neglected.  When comparing the wind velocity vectors in North and East direction, the vector in East direction indicates the least accuracy. As the quadrotor was operated in an environment where movement in East direction is constricted, it may have affected the accuracy of the wind velocity vector in East direction.

\section{Conclusion}
The findings of this study suggest that although the discrete form of quadrotor dynamics offers more precise wind velocity estimations compared to the continuous form in numerical simulations, when used with a real quadrotor, it indicated poor performance. The UKF estimator shows consistent results in both numerical simulations and the experimental setup. The EKF provided quicker estimations, however, the UKF, when used with the discrete dynamics, consistently yielded the most accurate results. This implies that UKF can replace widely used EKF in terms of accuracy. Despite using only cost-effective onboard sensors, the system was able to estimate wind velocity with satisfactory accuracy.  Consequently, the discrete form with the UKF can be applied to other precise estimation tasks beyond wind velocity estimation. Future research will aim to integrate wind velocity estimation in vertical direction, providing a full estimation of the surrounding wind condition of a quadrotor.


\bibliographystyle{IEEEtran}
\bibliography{citations}

\end{document}